\documentclass{article}
\usepackage{arxiv}
\usepackage[utf8]{inputenc} % allow utf-8 input
\usepackage[T1]{fontenc}    % use 8-bit T1 fonts
\usepackage{hyperref}       % hyperlinks
\usepackage{url}            % simple URL typesetting
\usepackage{booktabs}       % professional-quality tables
\usepackage{amsfonts}       % blackboard math symbols
\usepackage{nicefrac}       % compact symbols for 1/2, etc.
\usepackage{microtype}      % microtypography
\usepackage{graphicx}
\usepackage{amsmath}
\title{Meaningful uncertainties from deep neural network surrogates of large-scale numerical simulations}
\author{
 Gemma J. Anderson \\
 Lawrence Livermore National Laboratory \\
 Livermore, CA 94550 \\
  \texttt{anderson276@llnl.gov} \\
 \And
 Jim A. Gaffney \\
 Lawrence Livermore National Laboratory \\
 Livermore, CA 94550 \\
 \texttt{gaffney3@llnl.gov} \\
\And
Brian K. Spears \\
Lawrence Livermore National Laboratory \\
 Livermore, CA 94550 \\
 \texttt{spears9@llnl.gov} \\
 \And
 Peer-Timo Bremer \\
 Lawrence Livermore National Laboratory \\
 Livermore, CA 94550 \\
 \texttt{bremer5@llnl.gov} \\
 \And
 Rushil Anirudh \\
 Lawrence Livermore National Laboratory \\
 Livermore, CA 94550 \\
 \texttt{anirudh1@llnl.gov} \\
 \And
 Jayaraman J. Thiagarajan \\
 Lawrence Livermore National Laboratory \\
 Livermore, CA 94550 \\
 \texttt{jayaramanthi1@llnl.gov} \\
}

\begin{document}

\maketitle

\begin{abstract}
Large-scale numerical simulations are used across many scientific disciplines to facilitate experimental development and provide insights into underlying physical processes, but they come with a significant computational cost. Deep neural networks (DNNs) can serve as highly-accurate surrogate models, with the capacity to handle diverse datatypes, offering tremendous speed-ups for prediction and many other downstream tasks. An important use-case for these surrogates is the comparison between simulations and experiments; prediction uncertainty estimates are crucial for making such comparisons meaningful, yet standard DNNs do not provide them. In this work we define the fundamental requirements for a DNN to be useful for scientific applications, and demonstrate a general variational inference approach to equip predictions of scalar and image data from a DNN surrogate model trained on inertial confinement fusion simulations with calibrated Bayesian uncertainties. Critically, these uncertainties are interpretable, meaningful and preserve physics-correlations in the predicted quantities.
\end{abstract}


\section*{Introduction}
Numerical simulations are an invaluable tool to gain deeper insight into complex physical phenomena. They are used to improve our physical understanding, and can often help inform the design of future experiments. Unfortunately, they are usually very computationally expensive to run, prohibiting some state-of-the-art analysis that could push experimental design and scientific insight forward.

 %It is common in many scientific areas for simulations to have high-dimensional input parameter spaces and diverse output spaces, consisting of multiple modes, e.g. scalars {\it and} images. In order
%To capture the extremely complex non-linear interactions in the simulations, requires them to be run at high resolutions. Their significant computational expense often 
%resulting in a significant computational expense.
To alleviate the computational burden, a set of pre-run simulations can be used to train a surrogate model or ``emulator'' that maps simulation inputs to outputs, allowing a rapid approximation to the simulator for any given input parameter configuration.  Deep neural networks (DNNs) often outperform other machine-learning approaches due to their ability to capture highly non-linear dynamics in the simulations. It is also common for numerical simulations to output {\it multimodal} data (e.g. scalars {\it and} images) which can be handled directly with DNNs, avoiding the need for manual featurization of rich data structures which may discard important information. In fact, recent work using an autoencoder architecture has shown how DNNs can combine scalar and image outputs into a physically-meaningful, lower dimensional, representation of the multimodal data (referred to as a {\it latent space}). The auto-encoding network enables the model to capture important correlations between the data~\cite{hinton2006reducing} making this an important development for scientific deep learning. 

When using a surrogate model, uncertainty in its predictions must be quantified. A key drawback of standard DNNs is that they do not supply their predictions with a measure of uncertainty, making their use problematic in the sciences (and many other disciplines), where prediction uncertainties are typically as important as the prediction itself~\cite{Ghahramani_2015}. We often want to know to what extent we can trust our model in order to make important decisions. For example, in inertial confinement fusion (ICF) research, a tremendous effort is made to calibrate experimental diagnostics, but if we compare with a surrogate model (DNN or otherwise) that has missing or incorrect uncertainties this can result in misleading inferences. 

An important contribution to DNN prediction uncertainty is {\it epistemic}, or model form, uncertainty. In essence, DNNs are under-constrained, meaning that it is often possible to train many different DNN models that % new
give comparable matches to training data (as measured by the mean-squared-error or a similar metric). One method of capturing this uncertainty is through
%fit the data well. 
Bayesian DNNs, which capture uncertainty in their predictions by treating the weights in the network as probabilistic~\cite{mackay1992practical,neal2012bayesian}. The assumption is that there is one ``correct'' model, but it cannot be identified due to insufficient data. Instead of selecting a single model (in this case a DNN with a particular configuration of weights and biases) to describe the data, {\textit all} possible models are considered, each weighted by their posterior probability of them being the correct model. In essence, it represents a Bayesian Model Average~\cite{Wilson_2020}. 
Bayesian DNNs are very computationally intensive~\cite{Gal_2016b}, but faster variational approximations exist, e.g.~\cite{graves2011practical,blundell2015weight}. A convenient approximation to a full Bayesian DNN that requires minimal modification to an already existing DNN architecture is Monte Carlo (MC) dropout~\cite{Gal_2016}. Dropout, a technique originally introduced to reduce overfitting in DNNs~\cite{srivastava2014dropout}, ``drops'' nodes in the network at random during the training phase, keeping only a certain fraction of them specified by a free parameter known as the {\it keep-rate}, reverting to using the entire network when making a prediction. The key insight of ~\cite{Gal_2016} was that by sampling from the DNN with dropout still activated at prediction time (instead of reverting to using the entire network) gives a Bayesian approximation to the epistemic uncertainty in the DNN. Instead of yielding a point estimate, the DNN returns a {\it predictive posterior distribution} on the predictions. MC dropout has been used to obtain epistemic uncertainty estimates from DNNs in many areas of science as far ranging as cosmology~\cite{Perreault_Levasseur_2017}, medical imaging~\cite{Leibig_2017}, chemistry~\cite{Janet_2017}, and many more. There is a significant problem, however; the magnitude of prediction uncertainties from dropout DNNs is determined by user-specified hyperparameters for which no intuitive interpretation is available. 

In this work, we propose to fix this issue by enforcing a consistency requirement on the prediction uncertainties over a held-out validation dataset. A crucial criterion for probabilistic predictions is {\it calibration}, also sometimes referred to as {\it reliability}. This requires that the uncertainty in a prediction properly represents the probability that the prediction is correct, for example the 95\% confidence interval should contain the correct answer 95\% of the time~\cite{sanders1958evaluation,dawid1982well}. In our case, this is evaluated for the held-out validation set. Having properly calibrated prediction uncertainties is necessary for making reliable, trustworthy inferences, and for any subsequent decision-making. This criterion has its roots in probabilistic forecasting, but has also been recognized by the scientific community using deep learning for prediction~\cite{Perreault_Levasseur_2017}. Crucially, the epistemic uncertainties obtained from Bayesian and approximately Bayesian DNNs are not inherently calibrated due to their dependence on user-specified priors. In the case of MC dropout, prediction uncertainties in a single scalar output can be calibrated by tuning the dropout keep-rate~\cite{Perreault_Levasseur_2017}, however this becomes significantly more difficult for problems with multiple outputs and/or non-scalar data due to physically-driven correlations between outputs/pixels exist and must be preserved. In a recent paper, a method was introduced that automatically tunes the dropout keep-rate~\cite{gal2017concrete}. However, their results show that the improvements to the calibration of their uncertainties were marginal.

In this paper, we propose an approach for producing calibrated uncertainties from a DNN surrogate model trained on numerical simulations that simultaneously predicts multiple scalars and images. Our calibration approach leverages an autoencoder architecture, encoding the scalars and images into a joint lower dimensional latent space. We apply MC dropout and tune the dropout keep-rate such that we achieve calibrated uncertainties in the latent space. 
%By performing calibration in the latent space, we demonstrate that we can obtain calibrated uncertainties on both the scalar and image quantities, and 
A key benefit of our approach is that the meaningful correlations between the uncertainties of the different scalars, and image pixels are preserved.

\section{Surrogate models in ICF}
The National Ignition Facility (NIF) at Lawrence Livermore National Laboratory is home to the largest laser in the world. The laser's 192 beamlines heat and compress a tiny capsule containing hydrogen fuel causing it to implode, fusing the hydrogen nuclei together, a process known as inertial confinement fusion (ICF). The goal of the NIF is to achieve ``ignition'', where the fusion reactions become self-sustaining.
Radiation-hydrodynamic simulation codes such as HYDRA~\cite{Marinak_1998} are used throughout the experimental process, aiding in design, analysis and interpretation. HYDRA simulations of NIF experiments take target, laser and physics parameters as input, predicting the experimental outputs in the form of scalars, x0-ray and neutron images, and line-of-site dependant quantities. Many important tasks that are computationally infeasible using the raw simulations alone can be made possible with a surrogate model.
For example, previous experimental laser ``shots'' can be used to constrain simulation inputs thereby improving our understanding of the underlying physics~\cite{Gaffney_2019,Humbird_2019}, they can be used to correct the discrepancy between simulation and experiments~\cite{Kustowski_2020}, enable the discovery of promising new implosion shapes~\cite{Peterson_2017}, and many more.

For this work, we employ a HYDRA ensemble that consists of 92,630 2D simulations designed to simulate a particular class of inertial confinement fusion experiments on the National Ignition Facility 
known as `BigFoot' targets. The simulations span a 9D input parameter space encompassing known design parameters, unknown details of the drive applied to the target, and potential physics deficiencies in the code. The outputs of the simulations are post-processed to provide simulated experimental data, and like the experiments are multimodal: 28 key scalar diagnostics such as neutron yield, burn width and bang time, and X-ray images of the implosion. 

\begin{figure}%[tbhp]
\centering
\includegraphics[scale=0.35]{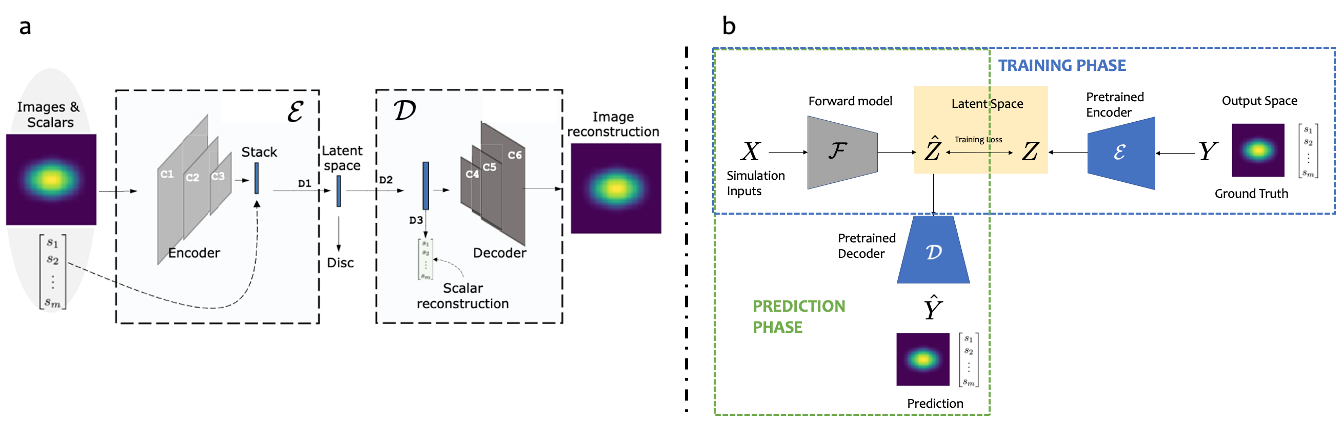}
\caption{MaCC network, adapted from~\cite{Anirudh_2020}.  In a) we show the autoencoder architecture, consisting of an encoder and decoder joined by a lower dimensional representation of the image and scalar data, known as a latent space. The dimension of the latent space was fixed at 32. In b) we show the forward model, a DNN with fully connected layers with non-linear ReLU activations, connecting simulation inputs to the latent space.}
\label{fig:auto}
\end{figure}

Recently, these data were used to develop a novel deep learning surrogate model known as the MaCC (manifold and cyclically consistent) model~\cite{Anirudh_2020}. MaCC was shown to make fast and accurate predictions of multimodal (scalar and image) ICF simulation outputs, $\bold{Y}= \{\bold{y}_i\}_{i=1}^{N}$, where $i$ refers to a particular simulation or ``sample'' and $N$ refers to the total number of samples. The MaCC model captures correlations in the multimodal data using an {\it autoencoder} that learns a lower dimensional latent space representation of the data, $\bold{Z}= \{\bold{z}_i\}_{i=1}^{N}$. More specifically, it consists of an encoder DNN mapping the simulation scalar and image outputs into a latent space $\cal{E}: Y \mapsto Z$ and a decoder DNN
$\cal{D}: Z \mapsto Y$. The encoder takes the scalar and image output from the ICF simulations and maps it into the latent space, which has its dimensionality tuned to be as small as possible while still providing nearly perfect reconstruction accuracy. The autoencoder architecture is shown in Figure~\ref{fig:auto}a. Given a reliable autoencoder, the MaCC model has a separate DNN that learns a mapping from a vector of simulation inputs, $\bold{X} = \{\bold{x}_i\}_{i=1}^{N}$, into the latent space $\cal{F}: X \mapsto Z$, shown in Figure~\ref{fig:auto}b. 

%footnote{This was the use case described in the paper, but the model can be used for other applications.}
%
Like the majority of DNN architectures, the original MaCC model is {\it deterministic}: given inputs $\bold{x}$, it will make a point prediction $\hat{\bold{y}}$, where ``$\wedge$'' denotes a prediction, with no measure of uncertainty. In the following work we equip the MaCC model with prediction uncertainties in order to enable reliable comparison with experimental data taken on the NIF.

\section{Uncertainties in DNNs}

There are two main sources of uncertainty in DNNs: {\it aleatoric } and {\it epistemic}. Aleatoric uncertainty arises when there is observational noise or {\it stochasticity} in the inputs to the DNN, for example in the case of experimentally-measured quantities, or images. This type of uncertainty cannot be reduced by collecting more data. In this work, the deep learning surrogate model is trained with pre-specified inputs and outputs that are from a deterministic simulation code therefore there is no stochasticity to consider in our set-up, however the generalization to include aleatoric uncertainty in simulation outputs (for example due to a Monte Carlo calculation) is straightforward using standard methods~\cite{kendall2017uncertainties} .
%[Most of this paragraph is relevant beyond DNNs; maybe it could be made more general].

Epistemic uncertainty, also known as model uncertainty, quantifies the DNN's lack of knowledge about the functional form of the simulation it is emulating. Epistemic uncertainty can be reduced with more simulation data. Bayesian DNNs capture this uncertainty by introducing a distribution of models and returning a {\it predictive posterior distribution}, marginalizing over the space of all plausible solutions consistent with the data. The posterior predictive distribution of a prediction ${\bf\hat{y}}$ for inputs ${\bf x}$ is given by
\begin{equation} \label{eq:ppd}
p({\bf\hat{y}}|{\bf x},{\bf X},{\bf Y}) = \int p ({\bf\hat{y}}|{\bf x},\omega) p(\omega | {\bf X},{\bf Y})d\omega,
\end{equation}

where $\omega$ denotes the DNN weights and biases. Eq.~(\ref{eq:ppd}) corresponds to a Bayesian model average over all possible explanations for the data. So instead of yielding a point estimate like standard DNNs, Bayesian DNNs give probabilistic outputs.
As well as equipping our predictions with uncertainty estimates, this approach also leads to improved accuracy compared to standard DNNs which are forced to choose a single model to describe the data, often not the most optimal one~\cite{Wilson_2020}.

Solving Eq.~(\ref{eq:ppd}) is challenging due to the very large number of parameters in a typical DNN, but it can be solved approximately. In this work, we implement MC dropout, which has the effect of approximately integrating over the DNN's weights, enabling us to equip the MaCC model with approximate Bayesian uncertainties. Critically for this work, the firm Bayesian foundation of MC dropout provides a meaningful statistical framework for our subsequent analysis.

We apply dropout before each fully connected layer in the $\cal{F}$ network with a pre-defined {\it keep-rate}, a parameter that determines the number of nodes to be dropped. The keep-rate is a parameter specifying the Bayesian prior on the DNN weights; since we have no prior context we propose to use the training data themselves to determine the keep-rate (a process sometimes known as {\it empirical Bayes}). For simplicity, we consider the autoencoder to be simply a non-linear preprocessing of the outputs and we do not consider any uncertainty arising from it. 
%We have investigated the inclusion of dropout in the autoencoder and found that the dropout splits between $\cal{F}$ and $\cal{D}$ with little effect on the total prediction uncertainty. 
We use the same network hyperparameters as in~\cite{Anirudh_2020}, except we do not include the cyclic loss term.

In order to obtain consistent results, we found it essential to use a training loss that includes uncertainty information. Without this, repeated training runs with the same keep-rate result in wildly different uncertainty models. Rather than the $L2$ loss used in previous work, we use the Gaussian likelihood loss given by
%Dropout measures epistemic uncertainty

\begin{equation} \label{eq:loss}
{\cal L} = \frac{1}{2\hat{\mathbf{\sigma}}^2} ||\mathbf{z} - \hat{\mathbf{\mu}}||^2 +\frac{1}{2}\text{log}(\hat{\sigma}^2),
\end{equation}

where $\mathbf{z}$ is ground truth in the latent space. This loss function aims to minimize both the prediction error and prediction variance. At each training iteration we generate 100 samples from the predictive posterior distribution, each using a different dropout mask. We chose 100 samples as this allowed for sufficient sampling of the distribution without being too computationally expensive to perform during the training process. The resulting distribution of predictions is used to compute the mean $\hat{\mu}$ and standard deviation $\hat{\sigma}$ above. Generating the approximate posterior predictive distributions of the latent space predictions is as simple as keeping dropout switched on at prediction time to generate 1000 samples from the network. The number of samples we generate is arbitrary, but since generating samples from a pretrained network is fast, we chose many more samples to ensure we sufficiently sample the posterior predictive distributions. 
The distributions are mapped using the decoder ${\cal D}$ network to obtain probabilistic predictions in the scalar and image outputs $\hat{y}$.
%The predictions from the $\bold{\cal{F}}$ network are probabilistic predictions in the latent space $\hat{z}$ that can be straightforwardly fed through the decoder ${\cal D}$ network to obtain probabilistic predictions in the scalar and image outputs $\hat{y}$.

\section{Calibrating uncertainty}
 In this section we define calibration, explain how calibration is measured, justify performing calibration in the latent space, demonstrate the importance of calibrating in the latent space in a toy problem, and present our results. 
 
\subsection{Calibration Definition}
 
 Our prediction uncertainties must be calibrated in order to be meaningfully compared with experiments. To be calibrated, if we a make a prediction with e.g. a $90\%$ confidence interval, then the ground truth should fall within that confidence interval $90\%$ of the time. We compute this for a held out test set $\{(x_{i},z_{i})\}_{i=1}^{N}$
 %use the trained DNN to make a prediction $\hat{z}_i$, by drawing 1000 dropout samples from the DNN. Subsequently, we compute the $p^{th}$ confidence interval for each test point and count the number of times this interval overlaps the ground truth value. Divide this by the total number of test points $N$ to get the predicted confidence $p(\hat{z})$. [I'm not sure about the wordy explanation. We didn't give a description of any of the other equations?] We can write this as
\begin{equation} \label{eq:calib}
p(\hat{z}) = \sum_{i=1}^N\frac{\mathbb{I} (\hat{z}_i^{\text{l}} < z_i < \hat{z}_i^{\text{u}})}{N},
%p(\hat{z}) = \sum_{i=1}^N\frac{\mathbb{I} (p^l(\hat{z}_i) \leq z_i \leq p^u(\hat{z}_i))}{N},
%\text{ECE} = \sum_{p \in \alpha} \bigg| p - \frac{1}{N} \sum_{i=1}^{N} \mathbbold{I}[\hat{p}_{z_i}^l  \leq z_{i}^{GT} \leq \hat{p}_{z_i}^u] \bigg | ,
\end{equation}
where $\mathbb{I}$ denotes the indicator function, the superscript $l$ and $u$ respectively correspond to the lower and upper bounds of the confidence interval. If calibrated, $p(\hat{z})$ should equal the confidence interval for which it was calculated, $p(z)$.

To visualize and diagnose the extent to which our DNN is calibrated we can plot the predicted confidence intervals $p(\hat{z})$ versus the observed confidence intervals $p(z)$, for a chosen number of confidence interval levels $p(z) \in [0,1]$, where here we choose $p(z)$ in $0.05$ increments, i.e. $p(z) = [0.05,0.1,0.15,\dots,0.95]$.

\subsection{Calibration Error Score}

A useful, succinct measure of calibration performance is provided by the calibration error score, given by

%Show calibration error plots as a function of keep rate in the latent space \\
%Show calibration curve in latent space for keep rate 0.98
\begin{equation}
\text{Calibration Error} = \sum w \big(p(z) - p(\hat{z})\big)^2,
\label{eq:ece}
\end{equation}
where $w$ denotes the relative weight of each confidence interval. Here we set $w \equiv 1$, giving equal weighting to all confidence intervals. This could be modified if it were determined that the confidence intervals had varying importance. Uncertainty estimates are considered {\it well-calibrated} if this score is close to zero.

\subsection{Latent Space Calibration}
The calibration error in Eq.~(\ref{eq:ece}) provides a metric for the quality of a given model similar to the Bayesian loss in Eq.~(\ref{eq:loss}). However this new metric is not differentiable with respect to the DNN weights $\omega$, meaning that it cannot be included explicitly in the training of the DNN. Instead, the keep-rate must be tuned outside of the DNN training loop~\cite{Perreault_Levasseur_2017}. %Recently,~\cite{gal2017concrete} introduced an extension to dropout known as concrete dropout which automatically tunes the dropout keep-rate for calibration. We were able to achieve well-calibrated uncertainties by tuning the keep-rate by hand, but for more complicated problems, this could prove to be a viable approach.

%introduces a prior distribution on the keep rate which allows it to be optimized simultaneously with the DNN weights [is that right?]. While the method presented here is more laborious, we have found that we can obtain superior performance. [We need to mention concrete dropout and this seems like a good place. However, we will probably need to include a comparison with CD in the supplemental material].

During training, the loss function  Eq.~(\ref{eq:loss}) tends to drive the variance of the prediction uncertainties (marginal on each output dimension) towards the local variance of the residuals. Therefore, calibrated uncertainties are automatically obtained when the distribution of residuals is Gaussian and uncorrelated between outputs. This is not the case for scientific problems; the various simulation outputs are strongly correlated by the underlying physics. For example, we may use both X-ray and neutron signals to diagnose ICF experiments which depend on the temperature of similar spatial regions of the plasma; similarly uncertainties between elements of non-scalar data like spectra and images are highly correlated simply by the diagnostics used to generate them. Therefore, attempting to calibrate uncertainties for the raw outputs in this way will give incorrect results. We address this problem by performing the calibration tuning in the latent space, which exists specifically to compress the highly correlated outputs into a small number of independent dimensions. While our autoencoder worked well in practice, including an explicit disentanglement objective (e.g. beta-VAE~\cite{higgins2016beta}) will ensure that the latent dimensions are more likely to be uncorrelated.

\subsection{A Toy Problem}
\begin{figure}%[tbhp]
\centering
\includegraphics[scale=0.8]{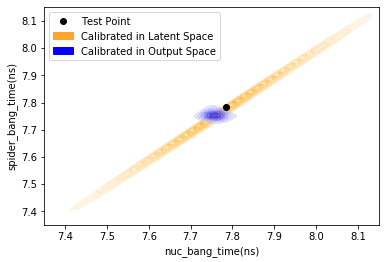}
\caption{Mean predicted value of a test point denoted in black shown as a function of two related scalar quantities, nuclear bang time and spider bang time, with uncertainty distribution where calibration was performed in output space (shown in blue) and where calibration was performed in latent space (shown in orange). This toy problem demonstrates that calibrating in latent space preserves important relationship between scalars.}
\label{fig:BT}
\end{figure}

\begin{figure}%[tbhp]
\centering
\includegraphics[scale=0.4]{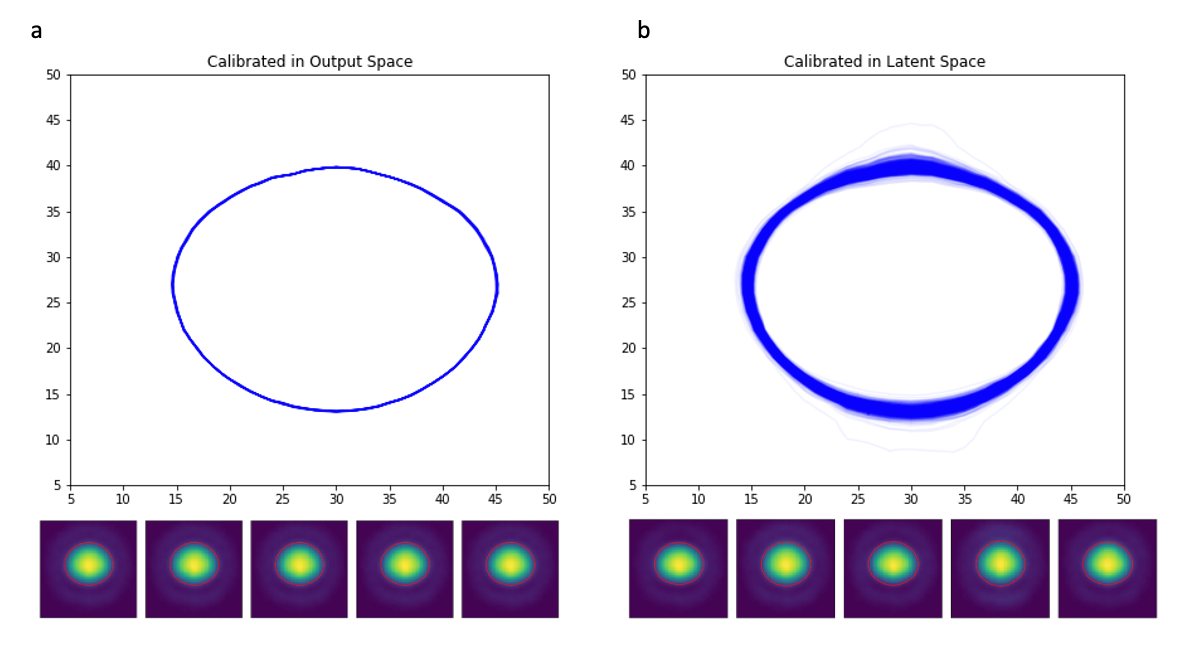}
\caption{$17\%$ contour for images with calibration performed in output space (left plot) and calibration performed in latent space (right plot). Example images with their $17\%$ contour plotted below}
\label{fig:17}
\end{figure}

We can demonstrate the importance of latent-space calibration using a relatively simple example. We train the MaCC model deterministically (i.e. no dropout), and use the distribution of the residuals as our uncertainty distribution. This gives us perfect calibration according to Eq.~(\ref{eq:ece}) but with no predictive capability and no statistical interpretation, both of which are also required to have meaningful uncertainties. We compare two cases: one where we use the residual distribution computed in the latent space as our uncertainty distribution and then propagate it through the decoder (calibrated in the latent space) and the other where we use the residual distribution computed in the output space as our uncertainty distribution (calibrated in the output space). We then compare the resulting uncertainties for a point prediction. Figure~\ref{fig:BT} shows predictions of two highly correlated scalar outputs, nuclear bang time and spider bang time, which measure the timing of the implosion using neutron and X-ray diagnostics respectively. Clearly the latent-space calibration (plotted in orange) captures the physical consistency requirement that the two diagnostics give related answers while the output-space calibration (plotted in blue) does not; in fact, portions of the blue distribution lie in regions of output space that were never valid outputs of the underlying physics simulations. Using such a model in a subsequent inference, for example, has the potential to generate physically unreasonable fits to the observed data.
Another advantage of calibrating in the latent space is that the calibration loss (\ref{eq:ece}), which only makes sense for uncorrelated scalar quantities, can still be used to generate calibrated uncertainties in non-scalar predictions. We show an example in Figure~\ref{fig:17}, where the same simplified uncertainty model is used to generate a set of predicted equatorial X-ray images for a single test point in input space. We plot a small set of predicted images, along with contours of constant brightness (17\% of peak) for each sampled image. Calibrating in the output space results in uncorrelated noise that gets averaged out and therefore results in no variation in the predicted 17\% contour, whereas calibrating in the latent space and then propagating to the output space results in increased, azimuthally-dependant uncertainties which are consistent with the sensitivities of the implosion shape to perturbations in our simulations.
\subsection{Results}
\begin{figure}[tbhp]
\centering
\includegraphics[scale=0.3]{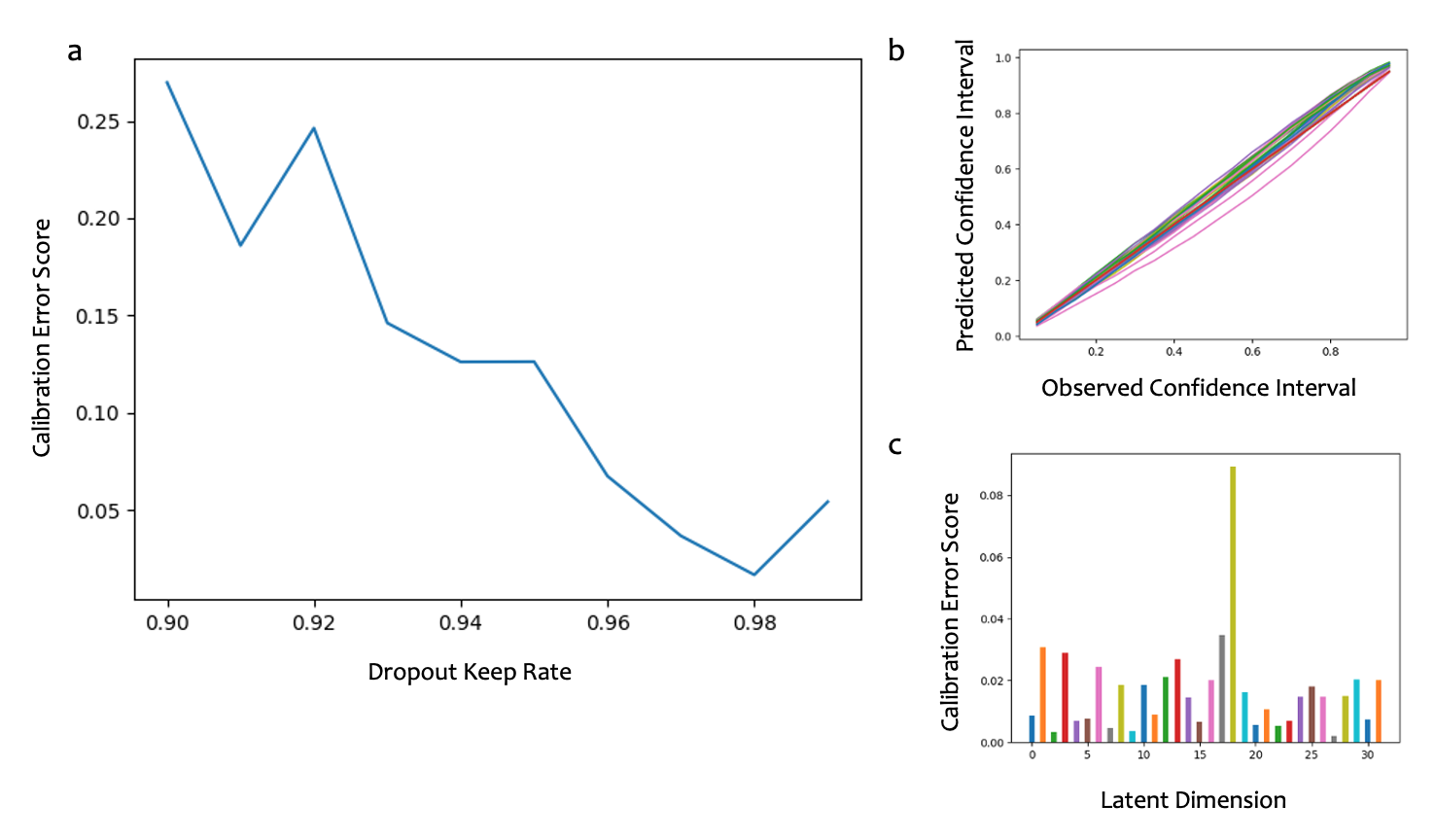}
\caption{(a) Calibration error scores as a function of keep-rate averaged over the 32 latent dimensions, (b) calibration curves for keep-rate 0.98 for each latent dimension, (c) calibration error scores for keep-rate 0.98 for the 32 latent dimensions}
% still need to fix up the plots once it's finalized
\label{fig:ece}
\end{figure}

The examples given above are instructive but do not provide a statistically meaningful, predicted uncertainty model. We now return to the fully Bayesian approach. We train a total of 10 uncertainty-equipped MaCC models with the dropout keep-rate varying from $[0.90-0.99]$ in increments of $0.01$. We plot the calibration error score, averaged over the latent dimensions in Figure~\ref{fig:ece}a. No dropout keep-rate we tried resulted in perfect calibration (a calibration error score of 0), but the model trained with a keep-rate of 0.98 was optimally calibrated, with very low calibration error scores for all of the latent dimensions. In Figure~\ref{fig:ece}b we show the calibration curves for all latent dimensions for this model demonstrating that the calibrated model gives excellent performance for all latent variables over all test confidence intervals (a model with perfectly calibrated uncertainties would lie along the diagonal line) - despite the fact that the Bayesian loss is only concerned with the variance along each dimension independently. This is a clear consequence of the normalizing effect of the latent space. We also show the calibration scores for each latent dimension from the optimally calibrated model in Figure~\ref{fig:ece}c. 
%According to Figure~\ref{fig:ece}c there is one outlier with significantly worse calibration; it turns out that this output [is there a reason? Not that I know of...].
%
%Show calibration error as a function of keep rate for each scalar to show that it is also optimally calibrated in the scalars  \\
%\subsection*{Meaningful predictions}
%\subsubsection*{Calibrated Scalars}
%Show an example of predictions for scalars (full predictive posteriors) and images (pixel-wise variance)
%
By leveraging an uncertainty-equipped deep learning architecture with a latent space, we have demonstrated for the first time that it is possible to obtain calibrated and meaningful uncertainties in scalar {\it and} image outputs. To the best of our knowledge, there are currently no existing methods that can provide calibrated prediction uncertainties on images that captures the physically-meaningful correlations between pixels.
%I think it's remarkable that this works with only a single dropout probability and a relatively simple modification to the loss. I would consider putting a summary paragraph in here that really hits people over the head with the result, and reinforces the message that the latent space is the key point.
\section{Uncertainty Interpolation}

A major challenge in using deep learning for scientific problems is the expense in creating training data. The multiphysics simulations used here, for example, represent hundreds of millions of CPU-hours on some of the largest HPC platforms in the world. There is therefore significant interest in answering the question: `how many simulations is enough?'. With our new calibrated uncertainties, we can answer that question.

\begin{figure}%[tbhp]
\centering
\includegraphics[scale=0.3]{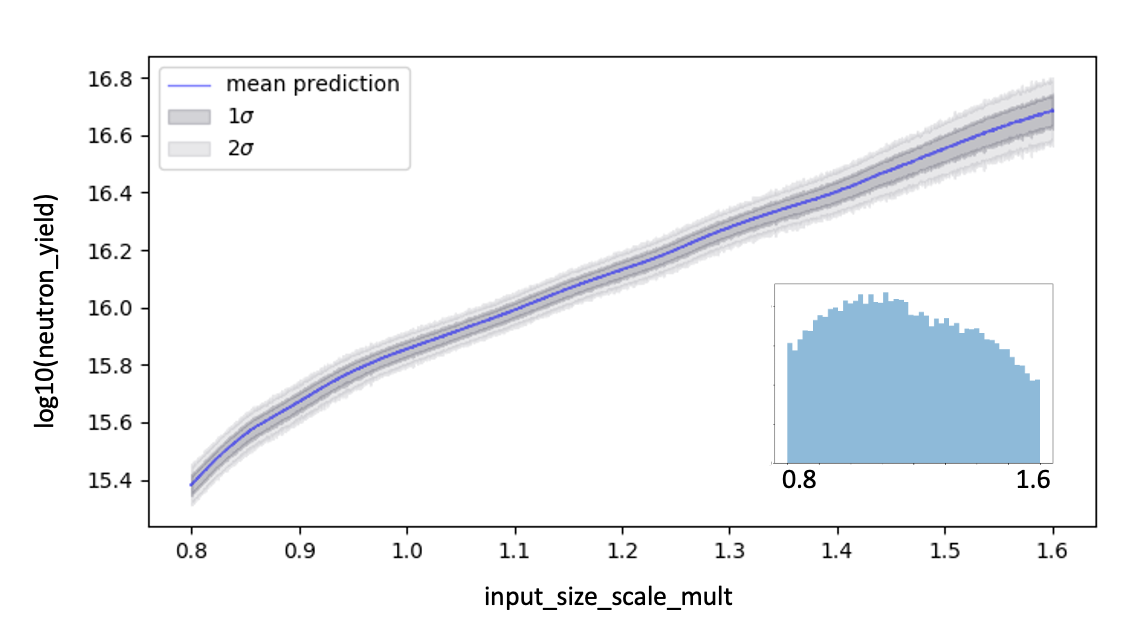}
\caption{Mean prediction of log neutron yield as a function of input size scale multiplier shown in blue, $1\sigma$ and $2\sigma$ error bounds shown by dark and light grey shading respectively. Inset shows the histogram of samples of input size scale multiplier over same range as main plot. Model has greater uncertainty in regions with fewer training samples.}
\label{fig:uncertainty}
\end{figure}

A key benefit of our approach is that areas of input space with fewer training samples will have a higher uncertainty on their predictions, since in those areas there are many more weight configurations that could explain the data. To demonstrate this, we set 8 of the 9 inputs parameters to fixed values and draw 1000 random samples for the 9th input parameter, called ``input size scale multiplier'' over the range it was trained on. Using our uncertainty-equipped MaCC model we make prediction with uncertainty of the log neutron yield (a key scalar performance metric in ICF) for this set of inputs, plotted in Figure~\ref{fig:uncertainty}. Our uncertainty-equipped MaCC model gives a decrease in uncertainty in regions where it  was given more training samples, as shown by the inset histogram. For example, the standard deviation of a prediction of the neutron yield when the input size scale multiplier parameter is between 1.0 - 1.1 is approximately half that of when the parameter is between 1.5 - 1.6. There are approximately 1.75 times more samples between the former and the latter ranges.
%[can we give a couple of examples? whats the density and uncertainty at the peak and edge of the distribution?]. 

This capability opens the door for an iterative approach to the building of scientific deep learning tools: generate a set of simulations, train a calibrated surrogate (typically requiring much less computer time than the simulations themselves), iterate until prediction uncertainties are small enough. These uncertainties can then be used as inputs to an adaptive sampling or active learning framework~\cite{lookman2019active}.
\section*{Discussion and Conclusion}
The uncertainty-equipped MaCC model that we present in this paper can be applied to many different scientific applications. We have shown in the case of large-scale numerical ICF simulations that it provides interpretable and meaningful uncertainties in predictions of scalars {\it and} images. The prediction uncertainties are calibrated and preserve important correlations between outputs, crucial for comparison with experiments. This stems from using a technique with a strong statistical foundation, namely MC dropout, and performing hyperparameter tuning based on calibration in the latent space. This approach can be applied to any model with the capability of encoding data via an autoencoder network. We also demonstrate that our uncertainty estimates can be used to ascertain where to perform more simulations.

\section*{Acknowledgements}
The authors would like to thank Andre Goncalves and Bogdan Kustowski for his valuable feedback on a draft of this work. This work was performed under the auspices of the U.S. Department of Energy by the Lawrence Livermore National Laboratory under Contract No. DE-AC52-07NA27344, Lawrence Livermore National Security, LLC. This document was prepared as an account of the work sponsored by an agency of the United States Government. Neither the United States Government nor Lawrence Livermore National Security, LLC, nor any of their employees makes any warranty, expressed or implied, or assumes any legal liability or responsibility for the accuracy, completeness, or usefulness of any information, apparatus, product, or process disclosed, or represents that its use would not infringe privately owned rights. Reference herein to any specific commercial product, process, or service by trade name, trademark, manufacturer, or otherwise does not necessarily constitute or imply its endorsement, recommendation, or favoring by the United States Government or Lawrence Livermore National Security, LLC. The views and opinions of the authors expressed herein do not necessarily state or reflect those of the United States Government or Lawrence Livermore National Security, LLC, and shall not be used for advertising or product endorsement purposes. This work was supported by LLNL Laboratory Directed Research and Development project 18-SI-002 and released with LLNL tracking number LLNL-JRNL-812258.

\bibliographystyle{unsrt}  
% Bibliography
\bibliography{references_new}

\end{document}